\documentclass[letterpaper, 10 pt, conference]{ieeeconf}  % Comment this line out if you need a4paper
\usepackage{graphicx}
\usepackage{tabularx}
\usepackage{multirow}
\usepackage{soul, color, xcolor}
\usepackage{algorithm}
\usepackage{algorithmic}
\usepackage{etoolbox}
\usepackage{makecell}

\makeatletter
\patchcmd{\@makecaption}
  {\scshape}
  {}
  {}
  {}
\makeatletter
\patchcmd{\@makecaption}
  {\\}
  {:\ }
  {}
  {}
\makeatother

\soulregister{\ref}7
\soulregister{\cite}7
\IEEEoverridecommandlockouts                              % This command is only needed if 
\title{\LARGE \bf
Simulated Mental Imagery for Robotic Task Planning
}

\author{Shijia Li, Tomas Kulvicius, Minija Tamosiunaite, and Florentin W\"org\"otter
\thanks{The research leading to these results has received funding from the German Science Foundation WO 388/16-1 and the European Commission, H2020-ICT-2018-20/H2020-ICT-2019-2, GA no.:871352, ReconCycle.}
\thanks{T. Kulvicius, M. Tamosiunaite and F. W\"org\"otter are with the Inst. of Physics 3, Dept. Computational Neuroscience, University of G\"ottingen, 37073 G\"ottingen, Germany, e-mail: shijia.li@phys.uni-goettingen.de.}% <-this % stops a space
\thanks{T. Kulvicius is also with University Medical Center G\"ottingen, Child and Adolescent Psychiatry and Psychotherapy, 37075 G\"ottingen, Germany.}
\thanks{M. Tamosiunaite is also with the Faculty of Computer Science, Vytautas Mangnus University, Kaunas, Lithuania.}
\thanks{Manuscript received April 19, 2005; revised August 26, 2015.}
}

\begin{document}

\maketitle
\thispagestyle{empty}
\pagestyle{empty}

%%%%%%%%%%%%%%%%%%%%%%%%%%%%%%%%%%%%%%%%%%%%%%%%%%%%%%%%%%%%%%%%%%%%%%%%%%%%%%%%
\begin{abstract}

Traditional AI-planning methods for task planning in robotics require a symbolically encoded domain description. While powerful in well-defined scenarios, as well as human-interpretable, setting this up requires substantial effort. Different from this, most everyday planning tasks are solved by humans intuitively, using mental imagery of the different planning steps. Here we suggest that the same approach can be used for robots, too, in cases which require only limited execution accuracy. In the current study, we propose a novel sub-symbolic method called Simulated Mental Imagery for Planning (SiMIP), which consists of perception, simulated action, success-checking and re-planning performed on 'imagined' images. We show that it is possible to implement mental imagery-based planning in an algorithmically sound way by combining regular convolutional neural networks and generative adversarial networks. With this method, the robot acquires the capability to use the initially existing scene to generate action plans without symbolic domain descriptions, while at the same time plans remain human-interpretable, different from deep reinforcement learning, which is an alternative sub-symbolic approach. We create a dataset from real scenes for a packing problem of having to correctly place different objects into different target slots. This way efficiency and success rate of this algorithm could be quantified.

\end{abstract}

\section{Introduction}
% ====================
% For Original Research Articles \citep{conference}, Clinical Trial Articles \citep{article}, and Technology Reports \citep{patent}, the introduction should be succinct, with no subheadings \citep{book}. For Case Reports the Introduction should include symptoms at presentation \citep{chapter}, physical exams and lab results \citep{dataset}.

Task planning is the process of generating an action sequence to achieve a certain goal. To do this with conventional AI-planning one needs to rigorously define symbolic structuring elements: the planning domain including planning operators, pre- and post-conditions, as well as search/planning algorithms \cite{c4,c5,c6}. While this is powerful in different complex scenarios, most every-day planning tasks are solved by humans without explicit structuring elements (even without pen\&paper).  Modern neural-network-based methods can predict the required action, given a scene, without any aforementioned symbolic pre-structuring \cite{schrittwieser2020mastering,hafner2020mastering}. However, the reasons for the decisions made by a neural networks usually remain opaque and interpretation by a human is impossible. Thus, networks elude explanations, which, however, might be important in human-robot cooperation tasks. Based on this need, we are suggesting a planning approach based on human-understandable entities: image segments, objects, and affordances, but no explicit domain descriptions.

%Often such plans of ours are based on multiple mental imagery steps. We imagine and mentally simulate a – usually short – sequence of actions with the seen objects and form mental images of the outcome of a planned action, which may be used as the start of the next imagined action. Often this does not even much enter our conscious though and planning depth is limited but we posit that a process that uses imagining for planning, acting, checking, and re-planning will many times also suffice for robots in everyday situations.

Our new method for task planning  called Simulated Mental Imagery for Planning (SiMIP) consists of the following components: perception, imagination of the action effect, success checking and (re)planning. This is similar to everyday human plans, comprising few steps only, being many times ad hoc, involving frequent success-checking and re-planning
\cite{c2}.  Note, however, that we abstract away from agent self-modeling (as e.g. in \cite{kwiatkowski2019task}) and only produce mental images of successive scenes. If one wants to extract parameters required for robotic execution, like locations of objects to be grasped or target locations of where to put the objects, one has to post-process the mental images showing scenes before the action and after the action.  In addition, we do not include actions of other agents  in our mental models (as e.g. in \cite{krichmar2019advantage}).

Extending affordance-based approaches,  which analyze one scene at a time \cite{c22}, we add to our architecture Generative Adversarial Networks (GANs) for simulated imagery of scenes following an action. %(and consequent affordance-based analysis therein). 
Given the impressive performance of GANs in realistic image generation \cite{c33, karras2019style}, one could potentially use them to envision outcomes of robot manipulation. However, when handling complex scenes, GANs tend to suffer from instabilities in learning. Also, when processing complex scenes in an end-to-end manner, network behaviour is hard to explain (e.g. see \cite{nair2019hierarchical}). Instead,  we suggest obtaining  future scenes by re-combinations on an object-by-object basis, with a GAN-based ”imagination” step for the completion of individual objects. This is reminiscent  of object-centric approaches that address scenes in object-by-object manner in latent space, (e.g. see \cite{veerapaneni2020entity,chang2022object}). However, we prefer to keep  the model explicit for achieving more stable training and performance.

As stated above, we use a simulated mental imagery process, which creates images of the outcome of an imagined action, then we use the imagined outcome as the input for the next imagined action, and so on. This way we can create a planning tree composed of images for which conventional search algorithms can be used to arrive at an image sequence that leads to the goal. 
While the tree remains sub-symbolic, due to the object-wise treatment of the imagined scenes, it can be readily post-processed into a symbolic representation required for robotic action. Stated in natural language, from the representations employed it is possible to deduce commands like that: ``pick an object with label A from the table top with center coordinate ($x_1$,$y_1$), and diameter  B cm and place it on an empty space with center coordinate ($x_2$,$y_2$)''. This, together with the obtained image trees, makes the approach  explainable to a human both in symbolic as well as in visual terms. 

We demonstrate our approach on a task of box packing, where we created and labeled a small data set for that.  As we keep the neural architectures simple (the aforementioned object-by-object attitude), comparatively small data sets suffice for training. Thus, one could also address a new task by preparing and labeling a new data at limited costs. 
%From a practical point of view, we address the domain of box-packing. 
A large domain of problems including packing, stacking and ordering (of table-tops, shelves) can be addressed this way.

The paper is structured as follows. In section II we discuss related work. Subsequently, an overview of our approach is presented in section III and implementation details are described in section IV. In Section V, we present experiments and results, and, finally, in section VI we provide a conclusion and outlook.

\section{RELATED WORK}

We will first briefly discuss classical symbolic, then neural-network-based sub-symbolic planning. We discuss  usage of  physical simulation in planning, in respect to mental imagery of future scenes used in our study.  Then we provide an overview of affordance recognition, focusing on aspects relevant to our framework. In the end we briefly review usage of neuro-symbolic representations in visual reasoning, which is also to some degree related to our approach.\\

\subsection{Symbolic Planning}

Classical planning techniques originating from STRIPS \cite{c4} are the usual choice for decision-making for robotic execution. They use a symbolic, logic-based notation compatible with human language that permits an intuitive domain specification \cite{c3}.  Contemporary planning techniques go a step forward and handle real world uncertainties using probabilistic approaches \cite{c7,c9, c11}. Despite the recent progress of such planning applied to robotics, these techniques are still subject to the symbolization problem mentioned before: all the relevant aspects for the successful execution of the robotic actions should be considered in the planning problem definition using scenario-specific domain descriptions.\\

To reduce hand-crafting, learning methods have been designed for aiding the domain definitions \cite{c8,ugur2015bottom, asai2018classical,ahmetoglu2022deepsym}. However, learning is not effort-free as data sets of pairs of pre- and post- conditions are required. In case of classical techniques, many constraints and problem pre-structuring is needed \cite{c8}. In case of deep learning approaches, most often latent space representations are used for obtaining ``symbols''. Experimentation how many symbols (i.e., latent variables) does one need is required \cite{ahmetoglu2022deepsym}, while any human-understandable meaning of these symbols can only be hand-assigned post hoc. Thus, symbolic representation learning, though possible, requires quite some additional design efforts. Generalization of the developed representations many times requires additional machinery, where objects and effects need to be assigned into classes, based on similarities in some feature space \cite{ugur2015bottom,james2022autonomous}, where the feature space is used for generalization afterwards. Thus, though promising, learning of planning operators remains relatively complex and, thus, is not frequently used in practice. \\

\subsection{Simulation}

Physical simulation is another way for future state prediction and simulation-based approaches for planning also exist. Fusion of simulation of sensing and robot control in virtual environments is an important development leading to the application of such techniques in robotics \cite{c17}. Planning of actions based on simulations has been done both in the realm of classical \cite{c19,c20} as well as deep-learning \cite{hafner2019learning} methods. To perform simulations, however, one needs robot- and object-models, as well as a full specification of the scenario. In industrial tasks, CAD models of parts and setups are usually available. However, this is usually not the case in everyday environments.  In this work, we are not concerned with industrial, high-precision robotic actions, but we are targeting the everyday-domain. There, most actions need only to be ``fairly'' accurate and, thus, one is not forced to simulate actions and their outcomes with the highest precision. Our method, thus, exploits mental simulation in the form of imagination of future scenes instead of physical simulation.\\

%This constraint underlies our approach similar to human actions, which are also not always super precise.

%\begin{figure}[htbp]
%  \centering
%  \includegraphics[width=0.45\textwidth]{examples_of_dataset.pdf}
%  \caption{Examples of the dataset}  \label{examples of dataset}
%\end{figure}
\begin{figure*}[htbp]
  \centering
  \includegraphics[width=0.7\textwidth]{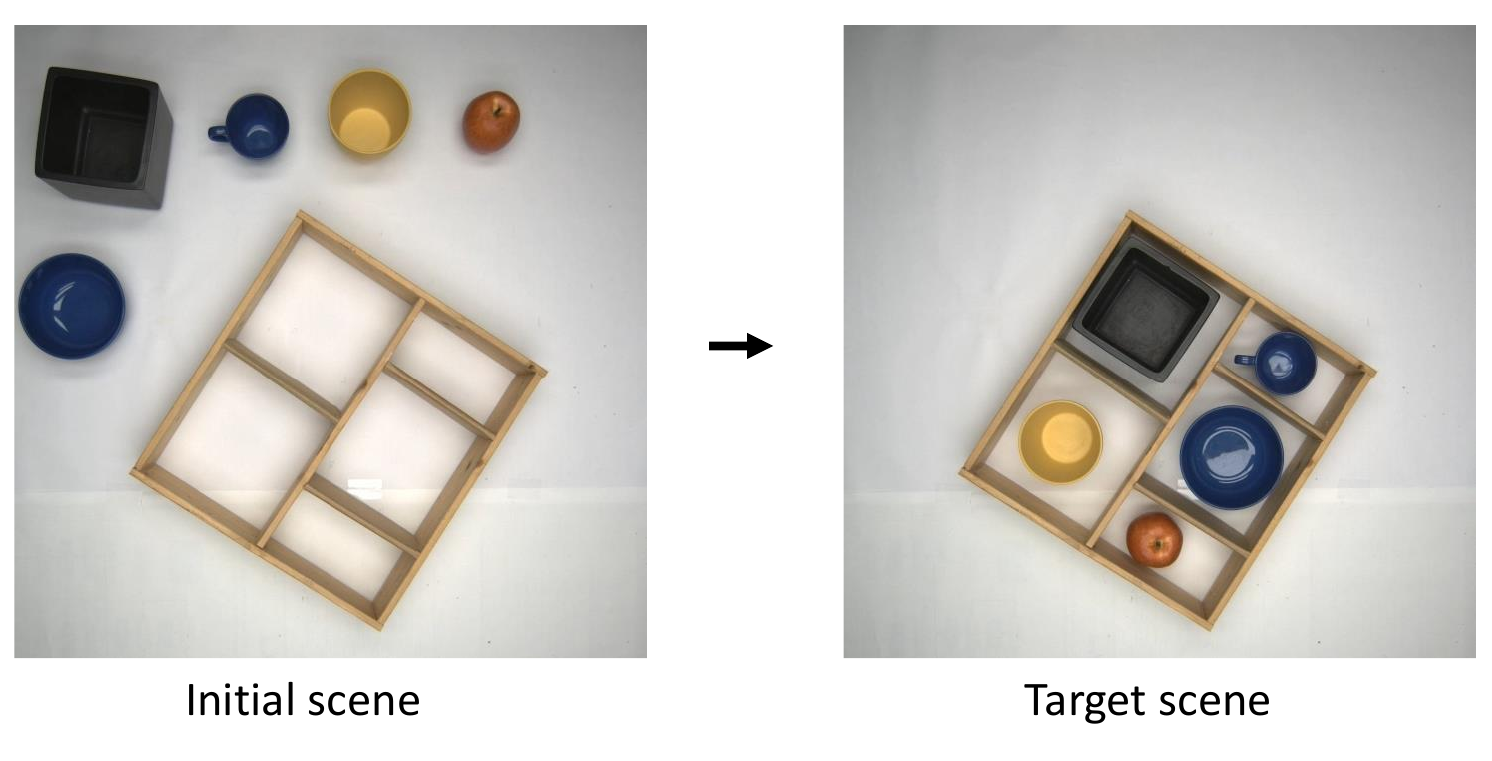}
  \caption{Task definition. The table top has to be ordered by putting all objects in the given box. The target is to leave no objects outside the box. Note, the ``target scene'' here is presented only for illustration purposes, as all other configurations, where there is  no object left outside the box, would be considered valid, too.}
  \label{goal}
\end{figure*}

\subsection{Sub-Symbolic Planning Using Neural Networks}

Deep reinforcement learning approaches allow learning action selection strategies in complicated worlds. Here explicit symbolic representations are not required, as actions can be deduced from the learned value function given the current scene (e.g. see \cite{schrittwieser2020mastering}, but  see also more citations, below). Such models are then capable of predicting future states, either at some level of abstraction, e.g. hidden/latent variables \cite{racaniere2017imagination,hafner2020mastering}, or as complete images \cite{ha2018world,kim2020learning}. Predicting future states helps training the models, as this way hypothetical future developments can be obtained. However, reinforcement learning requires probing very many consecutive states. Thus, such approaches, as for now, have been mainly developed  for computer games, where there are easy ways to register state-action sequences. When using imitation learning, which reduces data requirements, 3D simulated environments as well as real scenes can be addressed \cite{xu2019regression,lin2023mira}. Task and motion planning problem can be formulated and learned similar to reinforcement learning approaches \cite{driess2020deep}. Stereotyped tasks can be attained in real world experiments through long self-supervised experimentation by a robot \cite{ebert2018visual}, where this can be unavailable or too expensive for developing concrete applications. Different from all that, our approach does not require action sequences or pre- and post- condition pairs. Conventional approaches suffice here for learning of the following entities, which we need: object detection, object completion and affordance segmentation in the scene. These allow performing planning for us. 
%We do not address fine 3D pose related tasks as in \cite{lin2023mira}, which allows us to keep our methods for imagination simpler as compared to the aforementioned study. 

\vskip 0.5cm
\subsection{Affordance recognition}

The term “affordance” originates from cognitive psychology \cite{c21}. The set of affordances can be briefly described as the set of momentarily possible interactions between an agent and its environment. In robotics this term very often takes the meaning of: ``Which actions could a robot perform in a given situation (with some given objects)?''  The goal of affordance segmentation is to assign probabilities for a set of affordances to every location in an image.  A straightforward problem is trying to estimate affordances of whole objects \cite{c22, c23}. However, affordances can also be detected for multiple objects in the scene \cite{do2018affordancenet,lueddecke2019context}. Works exist predicting affordances resulting \textit{after} an action has been executed, this way aiding planning \cite{xu2021deep}.  Alternatively, here we will obtain future affordances through imagination of future scenes, thus pixel-wise affordance segmentation of scenes is enough for us.

\vskip 0.5cm
\subsection{Neuro-symbolic representations}
Related to planning are visual reasoning tasks, like visual question answering (VQA) \cite{yi2018neural, mao2019neuro, suarez2018ddrprog} which work through employing symbolic reasoning on images. These methods, similar to ours, include scene parsing modules, however, in addition, they heavily rely on NLP modules. We do not need NLP modules, as our aim is individual object manipulation, where object specificity beyond its affordances is not considered. Also related to our approach are video de-rendering tasks, where a latent representation is pushed towards an interpretable structure, by including a graphics engine into the decoder \cite{wu2017neural}. Other elaborate mechanisms exist to obtain symbolically-meaningful latent representations \cite{veerapaneni2020entity}.  We, however, do not go into the direction of interpretable latent representations, but rely on explicitly modeling individual objects in the scene as instances with affordances. Finally, graph neural networks may be applied for planning tasks, where geometric and symbolic information of a scene is supplied to the algorithm \cite{zhu2021hierarchical}. Different from all here mentioned algorithms, we avoid complicated network structures in order to avoid heavy demands on the amount of data required for training. We also avoid task pre-structured architectures, so that the application of the algorithm in a new situations is made easy. 

\begin{figure*}[htbp]
  \centering
  \includegraphics[width=0.85\textwidth]{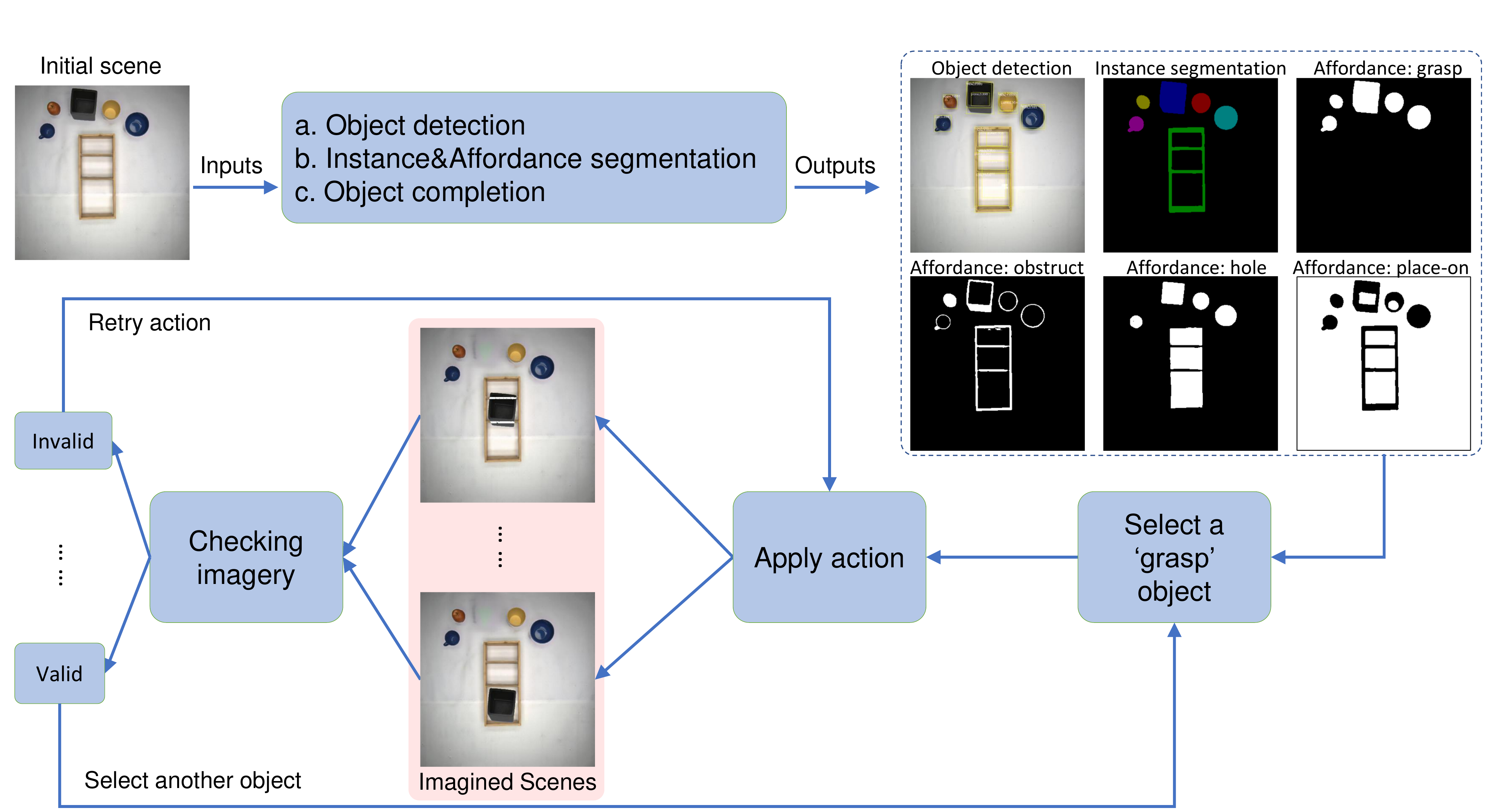}
  \caption{Flow diagram of our approach. Our system  contains two main parts: scene understanding and action planning. For scene understanding we use three deep networks, a) Object detection, b) Affordance\&Semantic segmentation, and c) Object completion. The details of the training and inference process can be seen in Fig.~\ref{structure}. Through the scene understanding part we can get the complete shape of the background and each individual object and its affordance class. Then, we can apply actions such as move and rotate to the object and use the information obtained from the affordance map to check whether the action is valid or not. If it is valid, we can perform the next action.}  \label{concept}
\end{figure*}

\section{ OVERVIEW}
\begin{figure*}[htbp]
  \centering
  \includegraphics[width=0.75\textwidth]{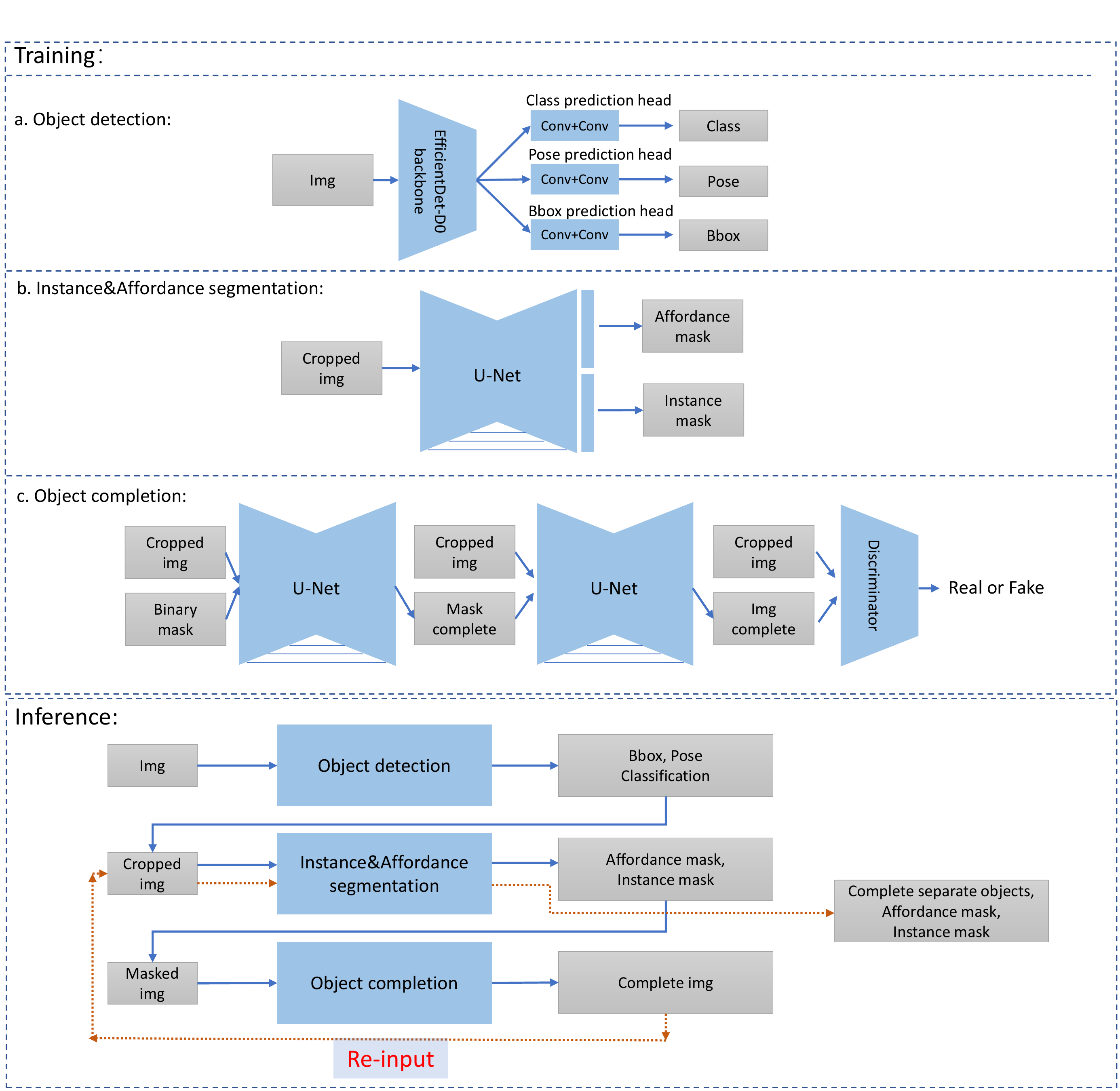}
  \caption{Training and inference of our model. In training: a) Object detection, b) Instance\& Affordance segmentation, c) Object completion (de-occlusion). In the training phase, we train the three models individually and then combine the obtained results in the inference phase. Note that after finishing the object completion (c, above), we need to do affordance segmentation (b, above) again, to get the complete object corresponding to the affordance classes (see red arrows). Bbox=bounding box. Details are explained in subsection \ref{impl_details}.  }
  \label{structure}
\end{figure*}

We are solving the task of ordering a desktop, where the system is presented with an initial scene (see Figure \ref{goal}) and the goal is to put the objects in the provided box, so that there are no objects remaining outside of the box. Thus, the algorithm is not provided with the target scene as such, but only with the condition that the table-top outside the box has to be free.  The box can be initially empty, as shown in the figure, or partially filled. Initial filling of the box may be incorrect, with too small objects occupying compartments required for putting in a bigger object. Furthermore, an initial scene with no objects outside of the box is also a valid scene, where we expect the answer from our algorithm to be that nothing needs be done.

In Fig.~\ref{concept}, the general workflow of our system is visualized, which we will describe next (for more details, see next section). We take as input an initial scene.  First, we perform object detection and pixel-level instance (object) segmentation. In addition to this, we also create an affordance map for the initial scene, which assigns to the scene pixel-level affordances. We then perform object completion (de-occlusion) using a Generative Adversarial Network (GAN). This allows us to split the whole scene into background and a set of complete individual objects. This is followed by pose estimation (not shown in the flow diagram).

Following that, we imaginarily-execute actions (i.e., generate post-action images), where we can choose from pick\&place, rotate, or flip vertically. After a post-action image was generated we perform a validity checking process  determining if an imagined action  has yielded a permissive result. 

Thus, in summary we have the workflow: a) Object detection, b) Instance\&Affordance segmentation, c) Object completion, d) Pose estimation, e) Application of the action f) Evaluation of the result. Parts a, b, c require deep network models, where the architectures for training and inference are given in Fig.~\ref{structure}. Details of implementation will be given in subsection \ref{impl_details} below.

\begin{figure*}[htbp]
  \centering
  \includegraphics[width=0.95\textwidth]{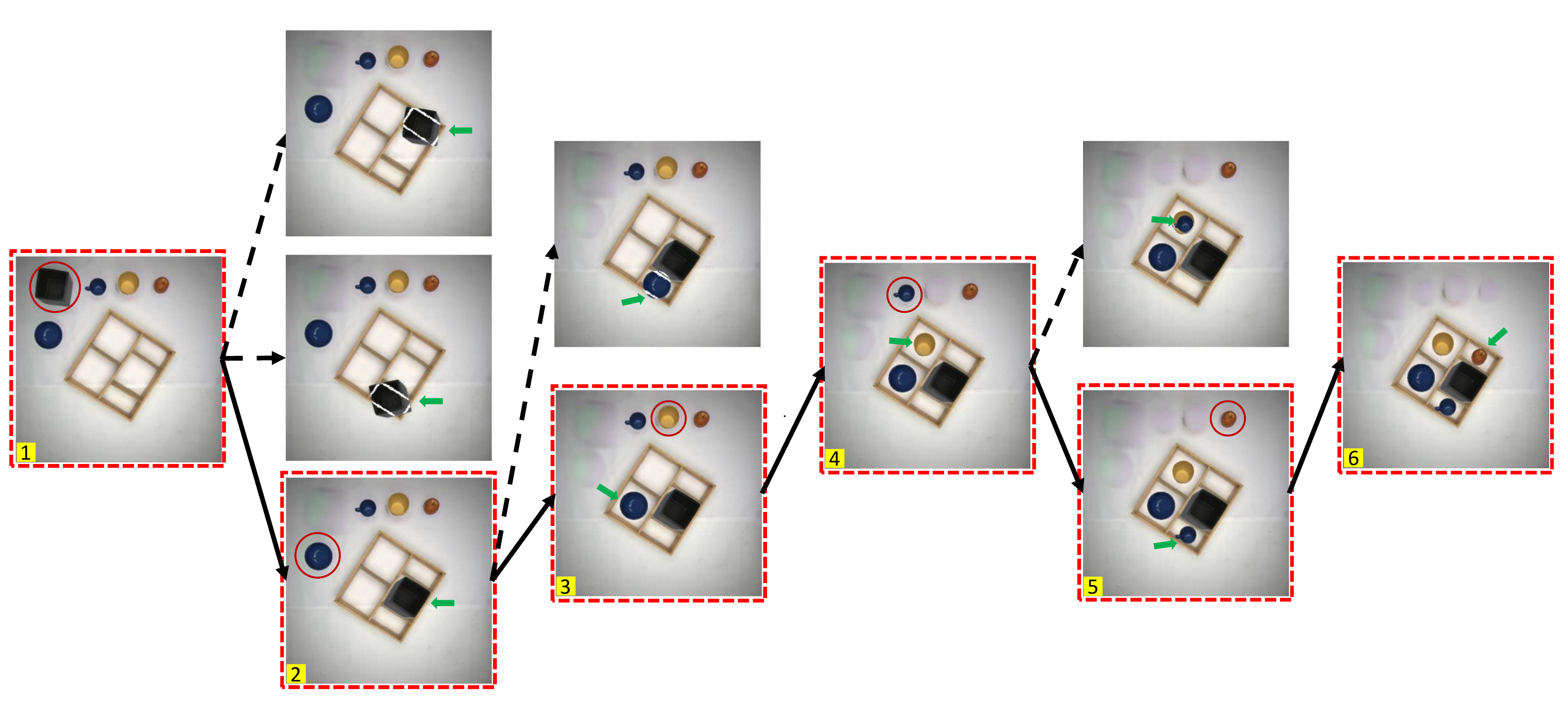}
  \caption{Demonstration of a planning tree. Each column represents an action step, the branches represent possible actions and each action is based on an imagined scene, where the previous action had been completed. The red dashed boxes mark the scenes indicating the valid planning sequence and are numbered consecutively (these numbers are used in Algorithm~\ref{alg}). Red circles indicate the objects on which the action is applied. The green pointer indicates where the object marked by the circle in the previous image has been placed.}
  \label{treelike}
\end{figure*}

\begin{algorithm*}
\caption{Translating a visual plan into a robotic-compatible symbolic plan with parameters. Entities and parameters used in translation are extracted alongside the imagination process.}
\label{alg}
\vskip 3mm
\underline{Variables:}\\
\textit{class\_label} \# determined for each object in each image, including box compartments \\
\textit{object\_name} \# of each individual object\\
\textit{image\_ref} \# reference to images in the valid plan sequence: image\_1 to image\_n \\
\textit{bounding\_box} \# parameters of the outer edges of the object in 3D\\
\textit{rotations\_angle} \# between the same object in two consecutive images

\vskip 3mm
\underline{Functions:} \\
\textbf{Bbox}(\textit{object\_name, image\_ref}) \#  Read out of a bounding box, given object name and image reference\\
\textbf{Grasp}(\textit{class\_label,bounding box}) \# for grasping of object, given its class label and bounding box\\
\textbf{Place\_at}(\textit{class\_label,bounding\_box}) \# for placing object at object of class label, given its bounding box\\
\textbf{Rotate}(\textit{object\_name,rotation\_angle}) \# for rotating of object using angle between two consecutive images\\
\textbf{Flip}(\textit{object\_name}) \# clockwise flip an object by $90^\circ$.

\vskip 3mm
\underline{Variable Assignment (for the example in Fig.~\ref{treelike}):}\\
\textit{class\_label}  = \{“can”, “bowl”, “cup”, “apple”,“compartment”\} \\
\textit{object\_name}= \{“black\_can”, “blue\_bowl”, “yellow\_cup”, “blue\_cup”, “red\_apple”, 

\hskip 4.5cm “compartment\_1”, “compartment\_2”,…, “compartment\_5”\} \\
\textit{image\_ref}=\{“image\_1”,\dots, “image\_6”\} \\
\textit{bounding\_box}=\{parameters of all bounding boxes\}  \\
\textit{rotation\_angel} = \{parameters of all rotation angles of all objects\}

\vskip 3mm
\underline{Plan (for the example in Fig.~\ref{treelike}):} \\
\textbf{Grasp}(“can”, \textbf{Bbox}(“black\_can”, “image\_1”)) \\
\textbf{Rotate}(“black\_can”,rotation\_angle) \\
\textbf{Place\_at}(“compartment”, \textbf{Bbox}(“compartment\_4”, “image\_1”)) \\
\textbf{Grasp}(“bowl”, \textbf{Bbox}(“blue\_bowl”,”image\_2”)) \\
\textbf{Place\_at}(“compartment”, \textbf{Bbox}(“compartment\_2”, “image\_2”)) \\
\textbf{Grasp}(“cup”, \textbf{Bbox}(“yellow\_cup”, “image\_3”)) \\
\textbf{Place\_at}(“compartment”, \textbf{Bbox}(“compartment\_1”, “image\_3”)) \\
\textbf{Grasp}(“blue\_cup”, \textbf{Bbox}(“blue\_cup”, “image\_4”)) \\
\textbf{Place\_at}(“compartment”, \textbf{Bbox}(“compartment\_5”, “image\_4”)) \\
\textbf{Grasp}(“apple”, \textbf{Bbox}(“red\_apple”, “image\_5”)) \\
\textbf{Place\_at}(“compartment”, \textbf{Bbox}(“compartment\_3”, “image\_5”))
\end{algorithm*}

\underline{Quantification:} We use a set of initial scenarios and create decision trees  based on imagined scenes (see Fig.~\ref{treelike})  and check validity of the scenes. All valid image sequences then represent valid plans, where pre- and post-conditions are implicitly encoded by the images. This way, we can quantify whether or not such a system shows degradation  along several planning steps, determining for different scenarios and manipulation sequences its actual usefulness for planning and execution. In Algorithm~\ref{alg} we show in a formal way how a symbolic plan for a robot can be extracted from the image-based plan shown in Figure~\ref{treelike}. For more details, see next section.

\section{IMPLEMENTATION}

\subsection{Data set}

The data to train and evaluate our proposed method is created from a real environment. We used a top-view camera positioned 100 cm above the center of the table and collect images with a resolution of $1024\times 768$ pixels.  Note that the usage of the top-view camera is not a restriction of this method. At the end of this study, we show that top views can be  generated by inverse perspective mapping. Hence, similar to human imagination processes, where we employ usually also some canonical "internal view" onto the imagined scene, also here the top view serves as the canonical perspective for our planning method.

The data set includes eleven different objects from seven classes: can, cup, plate, bowl, apple, box and cuboid, where cups and cuboids are two each and there are three different boxes. We use the following procedure for data collection: (a) we randomly place the box and some objects on a table; (b) we apply random actions by hand, changing position or orientation of one of the objects or the box and take a picture after each action has been accomplished. We repeat (a) and (b) multiple times. This way we collected 1196 scenes. Each scene contains at least one object with a unique pose and position. Afterwards, the scenes were labeled with instance and affordance annotations.  For instance annotation, the seven aforementioned object categories and four different affordance categories (grasp, place-on, obstruct, and hole; for description see Table~\ref{table:1})  were considered and extracted for all visible regions. It is important to note that our data set does not structure collected images into pairs: (image before  the action, image after the action) and does not include all possible goal configurations.
%\textcolor{red}{All data collection was conducted using a single overhead camera positioned approximately 100cm above the center of the table, providing a top-down perspective. Subsequently, we selected 1196 scenes from the recorded videos for segmentation class and affordance class annotations. It is important to note that our datasets does not include paired images for the starting and target actions, as our approach only requires a fuzzy objective.} For instance segmentation, seven different object categories (can, cup, plate, bowl, apple, box, and cuboid) and four different affordance categories (grasp, place-on, obstruct, and hole; for description see  Table~\ref{table:1}) were considered and extracted for all visible regions. In total, our data set consists of 1196 training samples, each manually annotated with categorical and affordance labels. 

%XXXXXXXXXXXXXXXX\textcolor{red}{To assess the performance of our approach, we conducted 5-fold cross-validation. All the networks are then trained and evaluated five times, each time using a different fold as the validation set and the remaining folds as the training set, with overall . By rotating the validation set across all five folds, we can obtain a comprehensive evaluation of the model's performance and ensure that the results are robust. Since the objects were randomly placed during data collection, each scene ensured the presence of at least one object with a unique pose and position. As our method operates on object-wise, the exact positions of objects have relatively less significance.} \\

\begin{table*}[!htb]
\centering
\caption{Description of the set of affordances used.}
\begin{tabularx}{0.8\textwidth} {|>{\hsize=.5\hsize\linewidth=\hsize}X|
>{\hsize=1.5\hsize\linewidth=\hsize}X|}
\hline
\textbf{Affordance} & \textbf{Description} \\
\hline
grasp & Areas that can be grasped to apply another action. \\ 
place-on & A surface, where objects can be placed. \\ 
obstruct & Areas, where objects are not allowed to be put. \\ 
hole & Hollow space in a solid object. \\ 
\hline
\end{tabularx}
\label{table:1}
\end{table*}

%\subsection{Experimental setup}
%In our experiments we have used a top view and a relatively small area, in which the objects reside, because this leads to only small projection-based deformations, which can be neglected. Processing of all steps has been performed on a local computer with an IntelCore i7-3770 CPU and a single NVIDIA 1080 Ti1080Ti GPU.

\newpage
\subsection{Network implementation details}
\label{impl_details}

Many of the approaches combined here represent standard methods, and will, thus, only be described briefly. Note, that neural networks for object detection, instance and affordance segmentation, as well object completion are first trained separately using our dedicated data set. Afterwards, the results from different networks are integrated to obtain the imagined planning tree, where we provide details of that integration, too.

For object detection, considering the size of the dataset, we used EfficientDet-D0 \cite{tan2020efficientdet} as the backbone network. During training, we apply horizontal flipping, scale jittering and randomly masking to perform data augmentation, and then we resized the images to 512$\times$512 pixels as input. We modified the output and added a pose classification head to predict whether the object is placed vertically or horizontally. The model is trained for 200 epochs with total batch size 4. We also used SGD optimizer with momentum 0.9 and reduced the initial learning rate 0.001 by factor 0.1 when the total loss has stopped improving after 3 epochs. The other parameters are same as in \cite{tan2020efficientdet} and the original loss functions are utilized.

For Instance\&Affordance segmentation, we used a U-net like architecture \cite{ronneberger2015u} and apply the same loss fuction as in \cite{lueddecke2019context}. Let $\rm C_sk$ denote the Convolution-BatchNorm-ReLU,  $\rm DC_sk$ double convolution layers (Convolution-BatchNorm-ReLU-Convolution-BatchNorm-ReLU), $\rm PC_sk$  partial convolution layers \cite{liu2018image} (PartialConvolution-BatchNorm-ReLU), k number of filters, and subscript s the stride. Furthermore, $\rm UP$ denotes upsampling layers, $\rm DN$  downsampling layers. Then the encoder is defined the following way: $\rm DC_132-DN-DC_164-DN-DC_1128-DN-DC_1256-DN-DC_1256$ and the decoder by: $\rm UP-DC_1512-UP-DC_1256-UP-DC_1128-UP-DC_164$. After the last decoder layer, two classification heads are applied to obtain four-dimensional output for affordance segmentation and two-dimensional output for semantic segmentation (background and main body). To work with the image with completed objects and for obtaining of the secondary segmentation mask, we cropped the image according to its axis-aligned bounding box and resized it to 256$\times$256 pixels. Combining all the outputs of the bounding box patches, we get the Instance\&Affordance segmentation for the original image. The model is trained for 400 epochs with the Adam optimizer and a learning rate 0.0001 with batch size 8. The binary cross-entropy loss is employed for classification.

For Object completion, we applied two U-net like architecture models PCNet-M and PCNet-C like in \cite{zhan2020self} as mask and image generator. For PCNet-M we used the same structure as used in segmentation and for PCNet-C we used 6 down-sampling layers encoder (PC$_2$64-PC$_2$128-PC$_2$256-PC$_2$512-PC$_2$512-PC$_2$512) and 6 up-sampling layers decoder (PC$_1$1024-UP-PC$_1$1024-UP-PC$_1$768-UP-PC$_1$384-UP-PC$_1$92-UP-PC$_1$67-UP). The last layer for the decoder has no BatchNorm and ReLU. For the discriminator, an SN-PatchGAN \cite{c33} is applied which uses 4 convolutional layers with spectral normalization (C$_2$64-C$_2$128-C$_2$256-C$_2$512) and one convolution map to  create a one-dimensional 70$\times$70 output. As in \cite{zhan2020self}, we also cropped each object according to its bounding box. The other parameters are same and the original loss functions are utilized. During training, the PCNet-M and PCNet-C are trained for 200 epochs and 800 epochs respectively with the Adam optimizer with learning rate 0.0001 and batch size 8. 

\begin{figure}[htbp]
  \centering
  \includegraphics[width=0.45\textwidth]{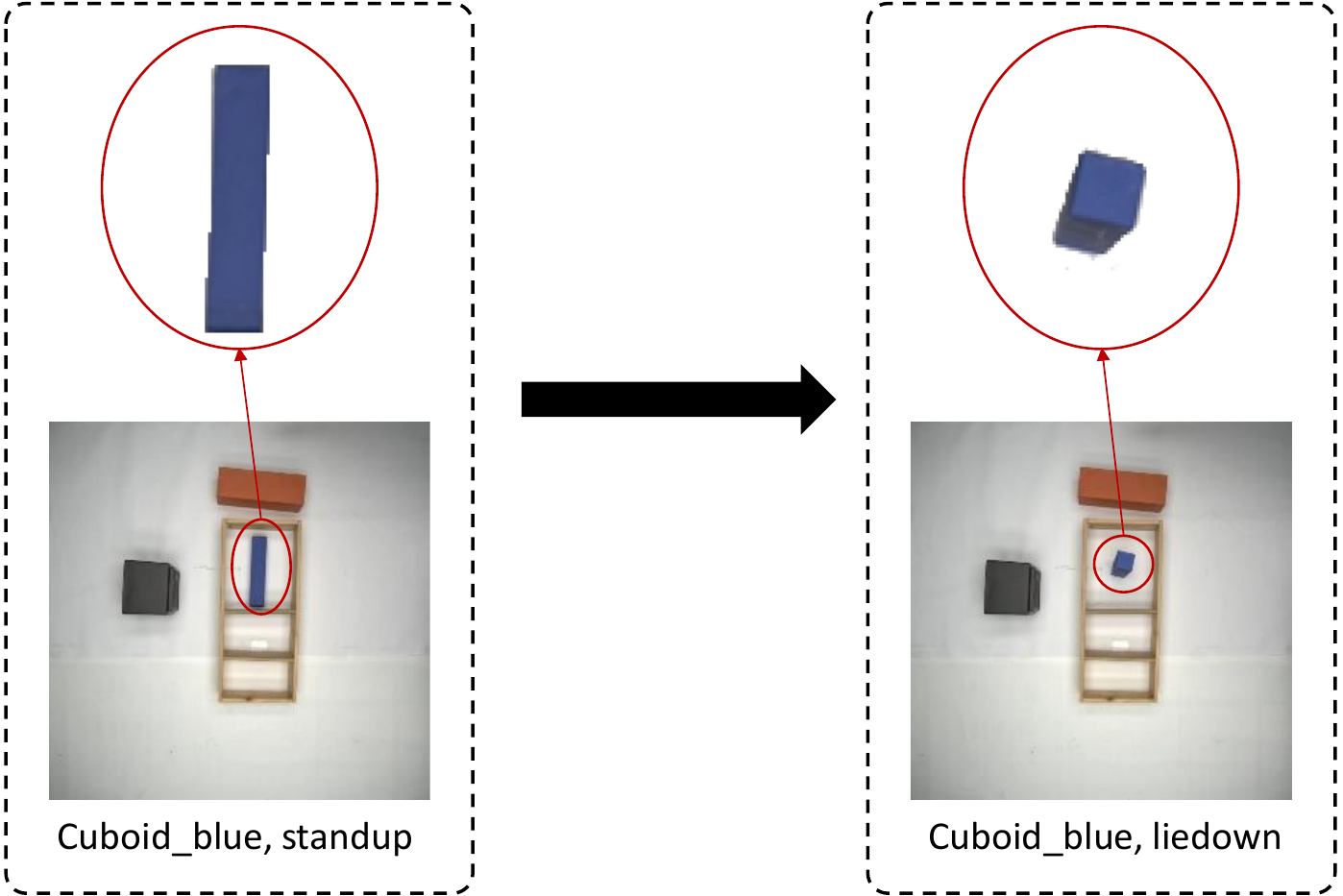}
  \caption{Example for pose mapping. We create a dictionary to store the horizontal and vertical pose of the blue cuboid. When we apply a flipping action on this object, we can lookup the dictionary and retrieve the corresponding pose}  \label{poses_mapping}
\end{figure}

\subsection{Pose estimation} 
The data collected in our experiments come from a top-down view of the RGB-camera image, which is appropriate for handling object movement and rotation in the horizontal direction. However, we also allow flipping of an object: for example, as shown in Fig.~\ref{poses_mapping}, cuboid will appear different when placed horizontally or vertically. Hence we need to predict the pose of the object, horizontal or vertical, together with object detection. Since each unique object in the experiment belongs to one category, we create a dictionary to store the horizontal and vertical poses of each object. The category and pose of the object are jointly used as primary keys, and the corresponding object's RGB image, instance segmentation map, and affordance map are saved as values. The horizontal or vertical pose of each category of objects is saved only once. When we need to flip an object, we can use this dictionary to get the flipped pose of the corresponding object.

\vskip 0.5cm
\subsection{Applying the action}
The actions, we apply, are pick~\&~place, rotate, and flip vertically. For pick~\&~place to be performed, the object has to have an affordance \textit{grasp} and the place where the object is placed shall have an affordance \textit{place-on}  or affordance \textit{hole}, however the \textit{obstruct} affordance shall be not present for those pixels. In the first instance, we do not check if the object is fitting on the area of affordance \textit{place-on} or \textit{hole} correctly, but see the next subsection ``Action validation'' where we solve this. For the action rotate, the object shall have affordance \textit{grasp}. Rotation is being performed in 15 deg. steps. For the action flip vertically, the object shall have affordance \textit{grasp} and the imagined action is performed by retrieving entries from the dictionary of horizontal vs. vertical poses, as was described above. The result of an imagined action is a post-action image. 

To obtain the post-action image we regard each object as a separate layer and then we use traditional image processing methods, such as cut-and-paste and rotation, to perform the movement and horizontal rotation of the object. We take the center of the object's bounding box as the origin when applying the action. For flipping objects, we need to replace the corresponding object layer with the flipped pose according to the dictionary. The object layers  are afterwards overlaid on a background layer to get the resulting image showing the result of applying the action. 

\vskip 0.5cm
\subsection{Action validation}
In the last step, having obtained images after action imagination, we check whether the action is valid or not. We require that the object is not placed in an area where the affordance is ‘obstruct’. Thus, the checking process is based on the affordance map. For this, we define conflict areas as the intersection of the 'obstruct' affordance with the manipulated object. We count the intersection pixels, where we set the threshold to 30 pixels. If the conflict area is less than 30 pixels, then we assume that the action is correct.

\vskip 0.5cm
\subsection{Formation of a planning tree}
For the planning tree we use a basic greedy search approach (depth-first-search) to generate a valid plan. For the initial scene, we first randomly select an object with a picking affordance from the set of objects standing outside the box. Then we attempt to position the object on a randomly selected placing affordance. For that, we perform a series of imaginary actions, including rotation and flipping and verify the image after each action until the object passes a validity check, which a conflict area of no more tan 30 pixels as described above. If success was not achieved by rotating the object in  15° steps either flipped or non-flipped, we proceed to choose another object from the ones standing outside the box.  If an object was successfully placed, we advance to the next planning step based on the image generated in the first planning step. Affordance-supported stacking here is also allowed. We terminate the process when there are no more objects outside or no action exists that passes the validity check. 

\vskip 0.5cm
\subsection{Parsing of symbolic entities}

In Algorithm~\ref{alg} we show an example of parsing of the valid visual plan, represented as an image sequence in Figure~\ref{treelike}, into symbolic planning entities with parameters required for robotic execution. Note, that entities used in the parsing process: class labels, bounding boxes, and the manipulated object sequence are directly obtained from the imagination process. Hence, for making a symbolic plan, it remains to collect those entities from the images and pass them to the corresponding robot action primitives. We provide the plan in an unwrapped form (instead of an algorithmic loop), to depict the full sequence of steps corresponding to the visual plan given in Figure~\ref{treelike}. Note, that this plan could be further processes (translated) into human readable sentences (not shown here) or – as an alternative – one could use automatic, neural network-based methods  (see e.g., \cite{dessi2023cross}) for image captioning to arrive also at a language-description of the images. However, the latter is much more demanding than the former, due to the fact that our system already provides many relevant entities and variables for sentence generation.

\begin{figure*}[htbp]
  \centering
  \includegraphics[width=0.5\textwidth]{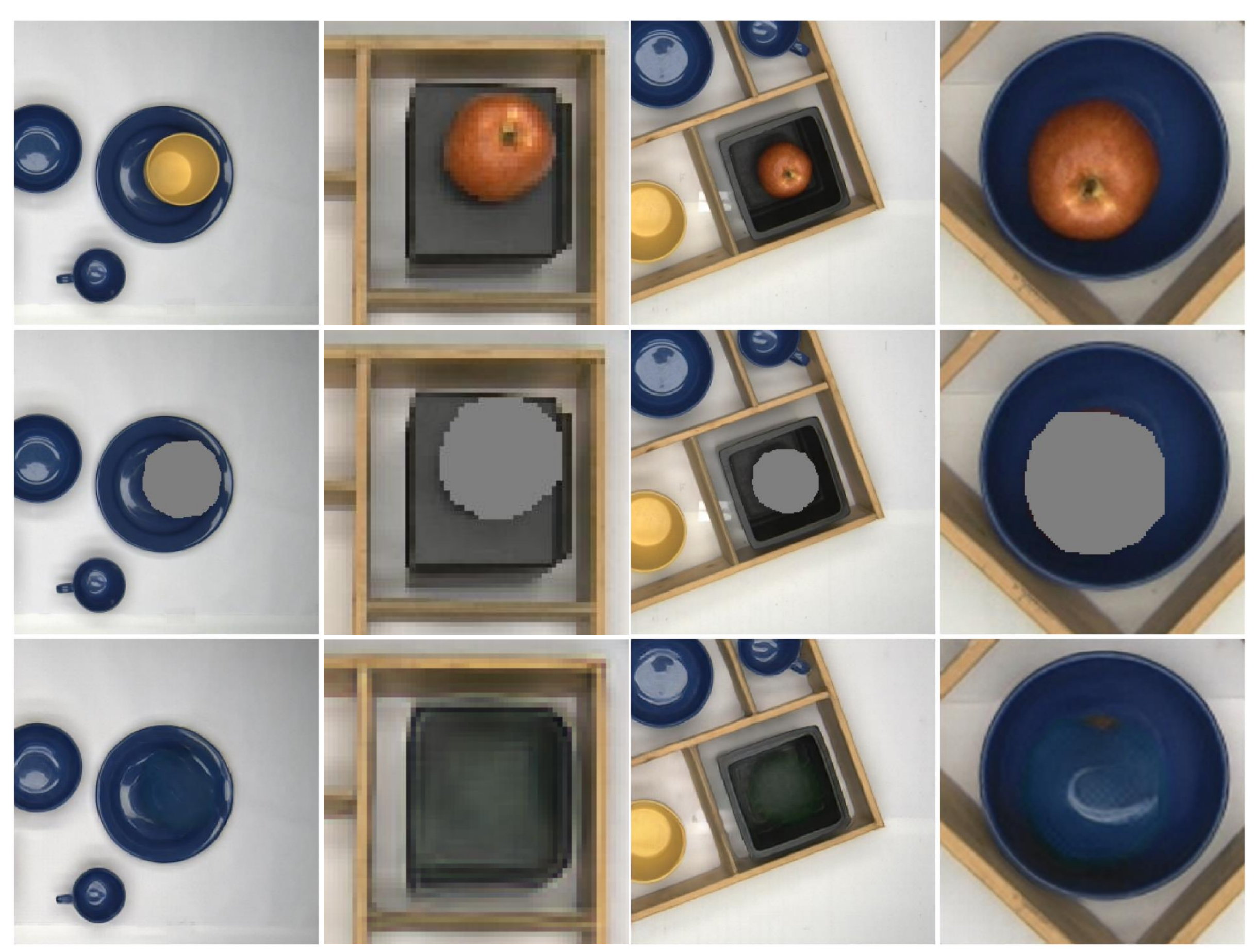}
  \caption{Object completion: qualitative examples. First row: image fragments from the test data set. Second row: mask of the obstructing object detected. Third row: completed object re-inserted into the scene.}  
  \label{completion}
\end{figure*}

%\vskip 1.5cm
\section{EXPERIMENTS \& RESULTS}

\begin{table*}[!htb]
\centering
\caption{The results for model components a, b, c (see Fig.~\ref{structure} first blue box on top). mAP -- mean average precision, mIoU -- mean intersection over union, L1 is the L1 norm; given as mean (SD) in 5-fold cross-validation.} 
\label{table:2}
\begin{tabularx}{0.95\textwidth} {|>{\hsize=0.5\hsize\linewidth=\hsize}X|
>{\hsize=0.5\hsize\linewidth=\hsize}X|}
\hline
\textbf{model}                    & \textbf{results: mean (SD)}              \\ \hline
Object detection: EfficientDet-D0 & mAP@0.5:0.95: 71.48\% (0.59\%)          \\ \hline
Instance segmentation: Unet       & mIoU: 94.18\% (0.36\%)              \\ \hline
Affordance segmentation: Unet     & mIoU: 91.70\% (0.89\%)               \\ \hline
Object completion:  PCNet-M\&PCNet-C    & L1: 0.0275 (8.37E-05) \ \  L2: 0.0028 (5.48E-05) \\ \hline
\end{tabularx}
\end{table*}

\begin{figure*}[htbp]
  \centering
  \includegraphics[width=1.0\textwidth]{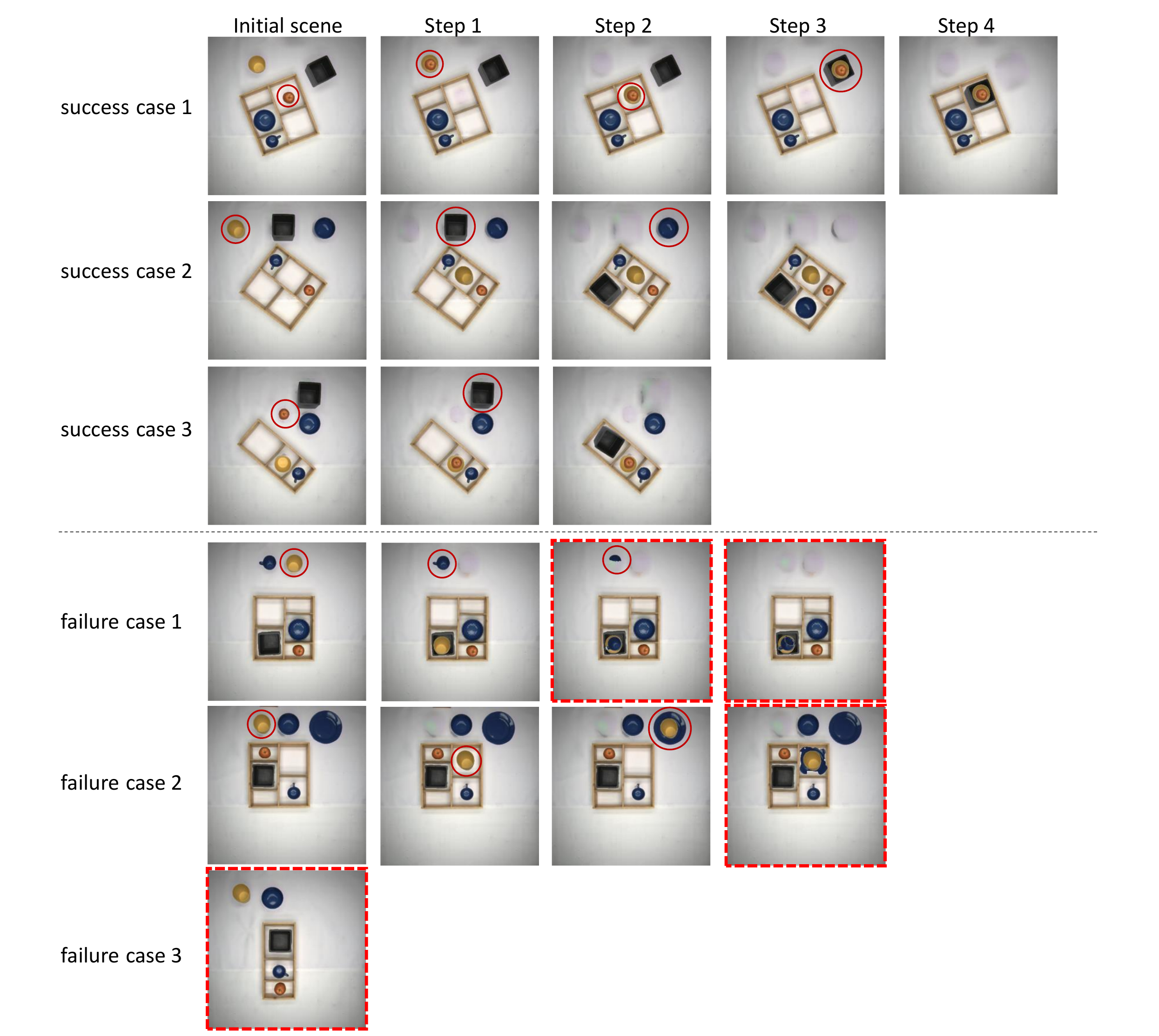}
  \caption{\tiny Examples of three successful and three failed plans. The first column represents the initial scene, each following column represents an individual imagined action step. Circles emphasize the objects for which the action is applied. Failed steps are marked with red dashed boxes. Explanation for individual cases: Success case 1: 1) Pick up \textit{red apple} \& place into \textit{yellow cup}. 2) Pickup \textit{yellow cup} (with \textit{red apple} inside) \& place into the \textit{compartment 4} of the box. 3) Pickup the \textit{yellow cup} (with the \textit{red apple} inside) \& place into \textit{black can}. 4) Pickup \textit{black can} (with \textit{yellow cup} and \textit{red apple} inside) \& place in the \textit{compartment 4} of the box. Success case 2: 1) Pick up \textit{yellow cup} and place into \textit{compartment 4}. 2) Pick up \textit{black can}, rotate 60 deg. and place into \textit{compartment 1} of the box. 3) Pick up \textit{blue bowl} and place into \textit{compartment 2} of the box. Success case 3: 1) Pick up \textit{red apple} and place into \textit{yellow cup}. 2) Pick up \textit{black can}, rotate 60 deg. and place into \textit{compartment 1}. In the failure cases, the red dashed box means an invalid step in a plan. In failure case 1, a part of the \textit{blue cup} is incorrectly identified as another cup. The same failure cause also happens in failure case 2, where a part of the \textit{blue plate} is identified as a can. In failure case 3, there were objects that could be packed, but no action was found in the search.}
  \label{plans}
\end{figure*}

As defined above, our task is defined as the need to organize a table top by packing objects into a box so that the table outside the box is empty.  The box has differently sized partitions and, similarly, objects have different sizes and shapes. 

First we will evaluate different system components: object detection, instance segmentation, affordance segmentation and object completion. Afterwords, we will evaluate the method as a whole, including ablation analysis. 
%For the definition of the action plan: Our task is to pack the objects into a box. In the beginning we need to put the objects one by one into the appropriate compartment according to their size, until all the objects on the desktop are put into the appropriate compartment, or there is no spare and appropriate compartment, the plan ends.

\begin{table*}[!htb]
\vspace*{2mm}
\centering
\caption{Success rates for planning cases with different plan lengths using a 5-fold cross-validation. Number of steps in plans denote factual number of steps in plans achieved by different methods. The results are reported as mean (SD). Best result in each column is emphasized in bold.}
\label{table:3}
    \resizebox{0.8\width}{!}{
    \begin{tabular}{|c|c|c|c|c|c|c|c|c|c|c|}
    \hline
        ~ & number of steps in plans  & 0 step & 1 step & 2 steps & 3 steps & 4 steps & 5 steps & 6 steps & 7 steps & total \\ \hline
        \multirow{2}*[-2ex]{our method} & total cases & \thead{28.6\\(3.44)} & \thead{56.2\\(6.94)} & \thead{34\\(2.83)} & \thead{47.4\\(2.97)} & \thead{39.2\\(4.97)} & \thead{19.8\\(5.02)} & \thead{7.2\\(1.3)} & \thead{7.6\\(0.89)} & \thead{240\\(0)} \\ \cline{2-11}
        ~ & success rate & \thead{\textbf{94.47\%}\\(1.60\%)} & \thead{\textbf{96.16\%}\\(1.24\%)} & \thead{\textbf{89.39\%}\\(1.63\%)} & \thead{\textbf{93.22\%}\\(1.14\%)} & \thead{\textbf{83.86\%}\\(4.47\%)} & \thead{\textbf{91.85\%}\\(1.93\%)} & \thead{\textbf{74.84\%}\\(6.31\%)} & \thead{\textbf{79.13\%}\\(6.42\%)} & \thead{\textbf{90.92\%}\\(1.19\%)} \\ \hline
        \multirow{2}*[-2ex]{\thead{our method \\ without object completion}} & total cases & \thead{26.6\\(3.65)} & \thead{60.6\\(6.27)} & \thead{32.4\\(3.71)} & \thead{45.2\\(2.49)} & \thead{40.2\\(6.65)} & \thead{21\\(4.06)} & \thead{7.4\\(1.14)} & \thead{6.6\\(2.19)} & \thead{240\\(0)} \\ \cline{2-11}
        ~ & success rate & \thead{87.57\%\\(4.38\%)} & \thead{95.01\%\\(2.15\%)} & \thead{80.93\%\\(2.73\%)} & \thead{89.77\%\\(2.82\%)} & \thead{77.25\%\\(3.26\%)} & \thead{91.52\%\\(0.86\%)} & \thead{68.02\%\\(7.57\%)} & \thead{52.06\%\\(10.49\%)} & \thead{86.00\%\\(0.96\%)} \\ \hline
        \multirow{2}*[-2ex]{baseline} & total cases & %\thead{13.8\\(1.79)} &
        \thead{--} &
        \thead{64.6\\(2.97)} & \thead{38.2\\(3.27)} & \thead{47.6\\(4.1)} & \thead{30.2\\(2.05)} & \thead{38.6\\(3.21)} & \thead{6.4\\(1.95)} & \thead{0.6\\(0.55)} & \thead{240\\(0)} \\ \cline{2-11}
        ~ & success rate &
        %\thead{100.00\%\\(0.00\%)} &
        \thead{--} &
        \thead{51.01\%\\(3.43\%)} & \thead{18.98\%\\(3.51\%)} & \thead{3.42\%\\(1.32\%)} & \thead{3.32\%\\(0.23\%)} & \thead{2.60\%\\(0.21\%)} & \thead{0.00\%\\(0.00\%)} & \thead{0.00\%\\(0.00\%)} & \thead{24.00\%\\(2.01\%)} \\ \hline
    \end{tabular}
    }
\end{table*}

\begin{table*}[!htb]
\vspace*{2mm}
\centering
\caption{Success rates for step by step analysis. The planning steps were performed for the testing dataset using a 5-fold cross-validation. The results are reported as mean(SD). Best result in each column is emphasized in bold.}
\label{table:4}
    \resizebox{0.9\width}{!}{
    \begin{tabular}{|c|c|c|c|c|c|c|c|c|c|}
    \hline
        ~ & n-th step of plans  & step 1 & step 2 & step 3 & step 4 & step 5 & step 6 & step 7 & total \\ \hline
        \multirow{2}*[-2ex]{our method} & total  steps & \thead{213\\(3.61)} & \thead{155.8\\(9.73)} & \thead{120.8\\(11.08)} & \thead{75.8\\(9.34)} & \thead{34.8\\(4.97)} & \thead{14.8\\(1.3)} & \thead{8.4\\(0.89)} & \thead{623.4\\(38.64)} \\ \cline{2-10}
        ~ & success rate & \thead{\textbf{97.84\%}\\(0.43\%)} & \thead{\textbf{97.17\%}\\(0.58\%)} & \thead{\textbf{97.31\%}\\(0.64\%)} & \thead{\textbf{95.33\%}\\(3.05\%)} & \thead{\textbf{95.45\%}\\(1.24\%)} & \thead{\textbf{95.99\%}\\(3.68\%)} & \thead{\textbf{93.00\%}\\(6.47\%)} & \thead{\textbf{97.03\%}\\(0.77\%)} \\ \hline
        \multirow{2}*[-2ex]{\thead{our method \\ without object completion}} & total  steps & \thead{215\\(2.92)} & \thead{152.8\\(7.22)} & \thead{120.4\\(10.21)} & \thead{75.2\\(8.79)} & \thead{35\\(4.24)} & \thead{14\\(2.35)} & \thead{6.6\\(2.19)} & \thead{619\\(29.33)} \\ \cline{2-10}
        ~ & success rate & \thead{95.90\%\\(0.86\%)} & \thead{94.08\%\\(1.12\%)} & \thead{97.10\%\\(1.24\%)} & \thead{93.56\%\\(2.59\%)} & \thead{92.07\%\\(1.79\%)} & \thead{88.64\%\\(3.07\%)} & \thead{62.86\%\\(16.63\%)} & \thead{94.71\%\\(0.76\%)} \\ \hline
        \multirow{2}*[-2ex]{baseline} & total  steps & \thead{226.2\\(1.79)} & \thead{161.6\\(3.85)} & \thead{123.4\\(4.39)} & \thead{75.8\\(6.22)} & \thead{45.6\\(4.98)} & \thead{7\\(2)} & \thead{0.6\\(0.55)} & \thead{640.2\\(18.79)} \\ \cline{2-10}
        ~ & success rate & \thead{56.60\%\\(2.93\%)} & \thead{27.97\%\\(1.80\%)} & \thead{19.96\%\\(2.07\%)} & \thead{8.51\%\\(1.48\%)} & \thead{3.00\%\\(0.84\%)} & \thead{0.00\%\\(0.00\%)} & \thead{0.00\%\\(0.00\%)} & \thead{32.12\%\\(1.25\%)} \\ \hline
    \end{tabular}
    }
\end{table*}

\vskip 0.5cm
\subsection{Evaluation of the system's components}

We evaluated the deep learning models used in our process on the test data set and the results are shown in Table~\ref{table:2}. Note that for the object completion task, we need to fill-in the occluded parts of objects, where we obtain small average losses L1 and L2 ( 0.0275 and 0.0028, respectively). As we are addressing the problem using a top-view, completion mostly addresses object stacks, however some small occlusions, occurring in case objects stand close together not directly under the camera, need this type of handling, too.  Since our data set is relatively small and the difference in object appearance between the training and test sets is not significant, these deep learning models perform well in our assigned task (see Figure \ref{completion}). Hence, these models build a solid foundation for the following task planning.

%Since the performance of  planning is strongly dependent on object detection and affordance segmentation results, our evaluation focuses on the success rate, as well as on the efficiency of the planning process.

\vskip 0.5cm
\subsection{Evaluation of the method}
We verified our method using 5-fold cross-validation. Scenes in the data sets differ in the number and location of objects. Our target is to place as many objects from outside the box as possible into the appropriate compartments in the box. To save computational resources, a depth-first search is used to find the complete plans, which are then checked whether they are valid. As many valid plans exist, it is costly to construct a ground-truth set for verification of validity (e.g. consider the need to account for all combinations of packing, including stacks of objects). Hence, we evaluated the obtained plans by eye.

Of the 240 test cases in each test set, there exist plans with zero up to seven packing steps. A ``0 step'' case corresponds to the situation where the box is fully packed and no planning steps are required. This is included to test the system's capability to recognize also such situations. Table~\ref{table:3} shows how many of these different cases had been successful. The grand average success rate across all cases was 90.92\%. As expected, the success rate deteriorates by 10-15\% for longer planning sequences, as both imagination and planning errors accumulate. 

In Table~\ref{table:4} we analyse all cases in a step-wise manner asking whether a plan step $n$ has been successful or not. On average 623 steps were performed across all 240 test cases in the plan-search process. The overall average success rate of one step is 97.03\%. Success rate deteriorates step-wise, however only by a couple percent from step 1 to step 7.  This demonstrates that  the imagination process used in our study degrades the images only minimally.

To identify the reasons of failed cases we analyzed the
causes of each failure. The failures can be attributed to
wrong object detection or inaccurate affordance segmentation results, which account for 45.37\% and 54.63\% of the
failure cases, respectively. Failures due to object completion can not be evaluated directly, thus, ablation study is made on that component, as described at the end of the section.

In Fig.~\ref{plans}, we show some successful and some failed plans. Three successful plans were able to complete our box packing task. The action sequences in those plans are described in the figure legend. In the failure cases, the red dashed box means an invalid step in a plan. In failure case 1, a part of the cup is incorrectly identified as another cup, which is caused by an inaccurate result of object detection. The same failure cause also happens in failure case 2, where a part of the plate is identified as a can, which in turn leads to a wrong action. In failure case 3, there were objects that could be packed, but no action was found in the search. This is, because none of the conflict areas calculated between the 'grasped' objects and all 'place-on' and 'hole' areas is smaller than the 30-pixel threshold value, which is caused by an inaccuracy of the result in affordance segmentation.

\vskip 0.5cm
\subsection{Comparison to baseline.} 
We performed a comparison to a baseline method where we randomly choose objects and placed them on random place-affordance locations for as many times (steps), as there were objects outside the box. For each test set the random placement was repeated four times to obtain more reliable averages. Results are shown in lines ``baseline'' in Tables~\ref{table:3}, \ref{table:4}. Note, that the baseline has a small advantage against our method, as it has information how many objects there are outside the box. This leads to deterministic 100\% performance in case there are no objects outside the box, thus, we consider this case as ``not applicable'' in Table~\ref{table:3}. Otherwise, the baseline performs substantially worse than our method, which is especially visible for longer plans. Note, the number of total steps in our method and in baseline method are different, as the baseline method uses a simplified procedure on decision how many steps are required.

\begin{table*}[!htb]
\centering
\caption{The results for initial scenes of successful and failure plans.} 
\label{table:5}
\begin{tabularx}{0.95\textwidth} {|>{\hsize=0.38\hsize\linewidth=\hsize}X|
>{\hsize=0.31\hsize\linewidth=\hsize}X|>{\hsize=0.31\hsize\linewidth=\hsize}X|}
\hline
\textbf{model}  & \textbf{successful plans:mean (SD)}  & \textbf{failure plans:mean (SD)} \\ \hline
Object detection: EfficientDet-D0 & mAP: 75.97\% (2.06\%)    & mAP: 70.96\% (0.19\%)      \\ \hline
Instance segmentation: Unet       & mIoU: 94.13\% (0.24\%)      & mIoU: 91.47\% (1.57\%)        \\ \hline
Affordance segmentation: Unet     & mIoU: 93.06\% (0.10\%)      & mIoU: 90.68\% (1.04\%)         \\ \hline

\end{tabularx}
\end{table*}

\vskip 0.5cm
\subsection{Ablation study.} 

Here we investigate the utility of different components. For the GAN-based component the results of the study are shown in Tables~\ref{table:3} and \ref{table:4}, lines ``without object completion''. In all cases the ablated version performs worse and the effect  becomes especially prominent in the last steps of the plan (see the last columns of Table~\ref{table:4}). This is expected, as with more imagination steps the need to reproduce object appearance grows. We did not make ablation study for other components of the method (e.g., object detection or affordance segmentation), as removal of those components disrupt operation of the framework completely. 

As we cannot completely exclude object detection, instance and affordance segmentation from the algorithm, we made those evaluations differently. We evaluate the influence of those components on the final result by calculating success measures of components for successful and failed plans separately, see Table~\ref{table:5}. One can see that the mean average precision (mAP) for object detection is 5\% smaller in failed cases, while mean intersection over union (mIoU) in instance and affordance segmentation is also a couple of percents smaller in failure as compared to success cases. This shows that there is a relation between success in the here analysed system components and the overall system performance.  

%It should be noted that the timing numbers discussed next are given only to provide some general information about performance levels. It is expected that search speed can be improved significantly when using faster hardware. Planning requires that the processes shown in the top panels in Fig.~\ref{concept} are performed once (for the start scene) and the processes at the bottom are repeated a certain number of times ("steps") until a valid plan is found. Of all 126 randomly generated starting configurations, we get plans where the minimum number of steps was 0 and the maximum was 7 steps, with an average of 2.7 steps required. Zero steps means that the initial scene is already in a packed state.  For the hardware we have used here, we get an average time to create a plan by searching of 8.175s and the average search time per iteration step is 3.003s. 

\begin{figure*}[htbp]
  \centering
  \includegraphics[width=0.95\textwidth]{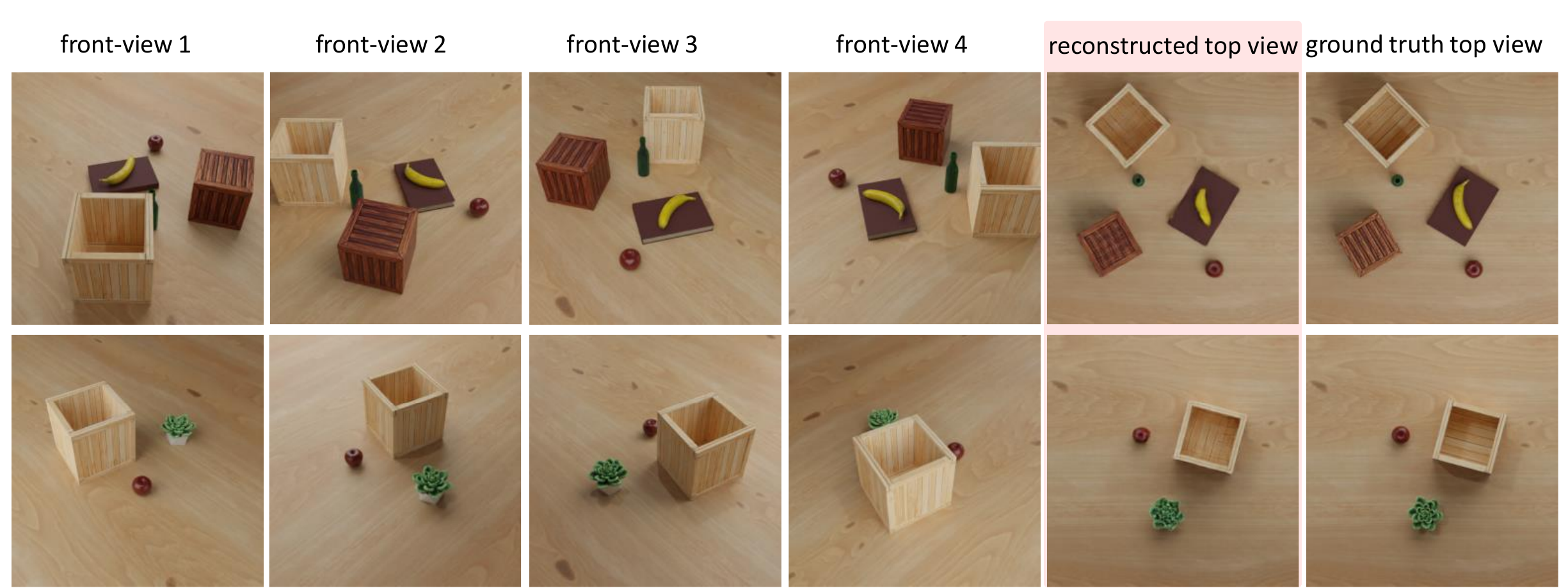}
  \caption{Example result of a simulated scene by applying inverse perspective mapping to create a top view from several side views.}  \label{front2top}
\end{figure*}

\section{DISCUSSION \& OUTLOOK}
We have presented a method for planning of packing and ordering tasks based on mental imagery, where a tree of imagined future scenes is created and the plan is represented as a sequence of images from such a tree. Unlike methods that predict entire future images in robot manipulation scenes end-to-end \cite{nair2019hierarchical}, our approach involves a scene parsing process, which brings the following advantages:
 \begin{itemize}
     \item Generative processes can be supported by comparatively small data sets;
     \item The parsed entities can be further used for definition of robotic actions.
 \end{itemize}

While successful operation of generative processes was proven in our ablation analysis, actual robotic action specification based on developed image sequences and robotic implementation will be addressed in future work.  

The approach supports explanation of the obtained plans to a human in a hybrid manner: symbolically, by using the labels of the parsed entities (see Algorithm~\ref{alg})  and at the sub-symbolic level by showing the human the pictures that were imagined by the system. E.g., by these pictures it is easy to see what would go wrong along those planning tree branches, which were not included into the valid plan. 

The developed system generalizes to different distributions of objects in the initial scenes and can achieve goal states not explicitly provided in the training data. However, the objects need to be learned for instance- and affordance segmentation as well as for generative object completion. The advantage of our method is that a relatively small data sets suffices and, thus, can be labeled with a concrete application in mind. Furthermore, due to the modularity of this system, each component within the system can be readily replaced with newly emerging state-of-the-art techniques.
%\textcolor{red}{Due to modularity, each component's method within the system can be readily replaced with newly emerging state-of-the-art techniques, providing flexibility and adaptability. In practical applications, pre-trained models can be employed and fine-tuned with limited data.} 

The current algorithm uses images obtained from a top view camera. This issue does not lead to restrictions, because one can recreate top view images using inverse perspective mapping methods as long as the ground plane is known. Fig.~\ref{front2top} shows how to generate top views from different camera perspectives. Here we created a simulated scene and placed four cameras at fixed positions around the scene for data collection. We first used inverse perspective mapping (IPM) to remap the images from four cameras into a preliminary orthographic projection based on the intrinsic and extrinsic camera parameters. Then we used a deep network (U-net) to further correct this distorted scene and to finally get a near optimal top view image. We used 2000 images for training and 200 images for testing. As this is not in the center of this study, we directly used top view cameras, instead, to generate a canonical view for all our experiments avoiding shape deformation, which might interfere with the planning process. However, if required, IPM pre-processing can be included into our algorithms without restrictions.

Concerning the generative process introduced in our study, we performed these on an object-by-object basis and this way achieved high performance, where future frames do not substantially deteriorate over time. Though generating full images of future scenes is in principle possible, and was addressed by several studies, e.g.~\cite{veerapaneni2020entity,nair2019hierarchical}, the obtained images are blurry (see Figures 7b in ~\cite{veerapaneni2020entity} and Figure 1 in ~\cite{nair2019hierarchical} ). In some own preliminary unpublished work, we were also attempting full image generation and saw the same deficiencies, too. Given that one needs anyhow individual object information for making robotic plans, applying object-by-object treatment of scenes, as now done in this study, is natural and reduces data requirements, while at the same time leading to satisfactory results.   

Clearly one cannot address very precise 3D fitting tasks for objects with complicated shapes with our approach and more specialized methods are required for that \cite{lin2023mira}. For generative approaches, more advanced methods like diffusion models \cite{rombach2022high} can be used. In general, existing works considering image-based foresight are mostly specialized, e.g. pouring \cite{wu2020can}, pushing, lifting and dragging \cite{ebert2018visual}, addressing only block-worlds, or rope manipulation \cite{wang2019learning}, closing a door and object pushing \cite{nair2019hierarchical}.  Here we show that for a packing, stacking, and ordering tasks one can simplify this by performing planning directly by visual imagination without pre/post-condition pairs for training in case of every-day accuracy requirements. In addition, from a practical perspective it is important, that for implementation of our system only deep-learning-based image analysis knowledge is needed,  while domain description or reinforcement learning knowledge is not required for that. 

%In addition to this, the here-resulting planning trees can provide a direct link to subsequent symbolic interpretation and reasoning as soon as one attaches labels to objects and actions. Some such tree could, for example, be parsed into brick-lying:placed:box:failed or brick-lying:flipped (upright):placed:box:success. From there on a next step can be to parse this into human language. Most other methods cited above, do not provide this sub-symbolic-to-symbolic link in a similar straight-forward way. While this is not part of the current study it offers a natural future extension.

Although our current work does not involve direct interaction with a real robot arm, for the implementation of the system on a robot, one can follow a similar approach as we did during data collection. A camera, capable of providing a top-down perspective, is required and it needs to be set up so that the robot does not occlude the scene when in its home position. The camera has to be synchronized with the robot so that it takes an image each time after the robot has accomplished an action and has returned to the home position. 
%\textcolor{blue}{In robotic scenario, in addition, a feedback loop can be implemented, where  measurements would be executed of how much a real situation differs from the predicted one in each step. In case of substantial differences re-planning can be implemented. Given errors accumulate in longer foresight processes as shown by our results, incorporation of a feedback could enhance the success rate of the plans. E.g. error like in Failure case 1, shown in Figure \ref{plans}, where part of a cup was left behind in forward imagination will be corrected and will not influence the final result.}

We also believe that incorporating feedback loops on a robot could enhance the success rate of the plans. We can assess the consistency between the actual scene after robot execution and the imagined scene to determine whether plan updates are necessary. If inconsistencies arise, we can choose to regenerate the plan, thereby improving the success rate of the task. E.g. errors like in Failure case 1, shown in Figure \ref{plans}, where part of a cup was left behind in forward imagination could this way be corrected and would then not influence the final result. Alternatively, we can also choose to regenerate the plan after each step, which allows for continuous updates of the overall plan. This approach essentially involves making predictions for each step individually and our experimental analysis above suggests that planning only one step yields high success rates. This will be the focus of our future work.

\section*{Conflict of Interest Statement}
% ---------------------------------------
%All financial, commercial or other relationships that might be perceived by the academic community as representing a potential conflict of interest must be disclosed. If no such relationship exists, authors will be asked to confirm the following statement: 

The authors declare that the research was conducted in the absence of any commercial or financial relationships that could be construed as a potential conflict of interest.

\section*{Author Contributions}
% -------------------------------
% The Author Contributions section is mandatory for all articles, 
% including articles by sole authors. If an appropriate statement is not 
% provided on submission, a standard one will be inserted during the 
% production process. The Author Contributions statement must describe 
% the contributions of individual authors referred to by their initials 
% and, in doing so, all authors agree to be accountable for the content 
% of the work. Please see  
% \href{http://home.frontiersin.org/about/author-guidelines#AuthorandContributors}{here} 
% for full authorship criteria.
S.L. performed data set generation, developed the methods and performed simulations and analyses. T.K. provided camera setup and data acquisition, M.T. and F.W. performed analyses and wrote manuscript.  

\section*{Funding}
% ----------------
% Details of all funding sources should be provided, including grant 
% numbers if applicable. Please ensure to add all necessary funding 
% information, as after publication this is no longer possible.
The research leading to these results has received funding from the German Science Foundation WO 388/16-1 and the European Commission, H2020-ICT-2018-20/H2020-ICT-2019-2, GA no.:871352, ReconCycle.

%\section*{Acknowledgments}
%We thank T. Lilienkamp and S. Luther for support with the \textsc{BOCF} model and for inspiring discussions about %spatio-temporal dynamics in excitable media.

\bibliographystyle{IEEEtran}
\bibliography{mentsim} % your .bib file

\end{document}